# A Closer Look at Spatiotemporal Convolutions for Action Recognition

Du Tran[1], Heng Wang[1], Lorenzo Torresani[1,2], Jamie Ray[1], Yann LeCun[1], Manohar Paluri[1]
[1]Facebook Research   [2]Dartmouth College
{trandu,hengwang,torresani,jamieray,yann,mano}@fb.com

## Abstract

*In this paper we discuss several forms of spatiotemporal convolutions for video analysis and study their effects on action recognition. Our motivation stems from the observation that 2D CNNs applied to individual frames of the video have remained solid performers in action recognition. In this work we empirically demonstrate the accuracy advantages of 3D CNNs over 2D CNNs within the framework of residual learning. Furthermore, we show that factorizing the 3D convolutional filters into separate spatial and temporal components yields significantly gains in accuracy. Our empirical study leads to the design of a new spatiotemporal convolutional block "R(2+1)D" which produces CNNs that achieve results comparable or superior to the state-of-the-art on Sports-1M, Kinetics, UCF101, and HMDB51.*

## 1. Introduction

Since the introduction of AlexNet [19], deep learning has galvanized the field of still-image recognition with a steady sequence of remarkable advances driven by insightful design innovations, such as smaller spatial filters [30], multi-scale convolutions [34], residual learning [13], and dense connections [14]. Conversely, it may be argued that the video domain has not yet witnessed its "AlexNet moment." While a deep network (I3D [4]) does currently hold the best results in action recognition, the margin of improvement over the best hand-crafted approach (iDT [38]) is not as impressive as in the case of image recognition. Furthermore, an image-based 2D CNN (ResNet-152 [25]) operating on individual frames of the video achieves performance remarkably close to the state-of-the-art on the challenging Sports-1M benchmark. This result is both surprising and frustrating, given that 2D CNNs are unable to model temporal information and motion patterns, which one would deem to be critical aspects for video analysis. Based on such results, one may postulate that temporal reasoning is not essential for accurate action recognition, because of the strong action class information already contained in the static frames of a sequence.

In this work, we challenge this view and revisit the role of temporal reasoning in action recognition by means of 3D CNNs, i.e., networks that perform 3D convolutions over the spatiotemporal video volume. While 3D CNNs have been widely explored in the setting of action recognition [15, 35, 36, 4], here we reconsider them within the framework of residual learning, which has been shown to be a powerful tool in the field of still-image recognition. We demonstrate that 3D ResNets significantly outperform 2D ResNets for the same depth when trained and evaluated on large-scale, challenging action recognition benchmarks such as Sports-1M [16] and Kinetics [17].

Inspired by these results, we introduce two new forms of spatiotemporal convolution that can be viewed as middle grounds between the extremes of 2D (spatial convolution) and full 3D. The first formulation is named mixed convolution (MC) and consists in employing 3D convolutions only in the early layers of the network, with 2D convolutions in the top layers. The rationale behind this design is that motion modeling is a low/mid-level operation that can be implemented via 3D convolutions in the early layers of a network, and spatial reasoning over these mid-level motion features (implemented by 2D convolutions in the top layers) leads to accurate action recognition. We show that MC ResNets yield roughly a 3-4% gain in clip-level accuracy over 2D ResNets of comparable capacity and they match the performance of 3D ResNets, which have 3 times as many parameters. The second spatiotemporal variant is a "(2+1)D" convolutional block, which explicitly factorizes 3D convolution into two separate and successive operations, a 2D spatial convolution and a 1D temporal convolution. What do we gain from such a decomposition? The first advantage is an additional nonlinear rectification between these two operations. This effectively doubles the number of nonlinearities compared to a network using full 3D convolutions for the same number of parameters, thus rendering the model capable of representing more complex functions. The second potential benefit is that the decomposition facilitates the optimization, yielding in practice both a lower training loss and a lower testing loss. In other words we find that, compared to full 3D filters where appearance and dy-namics are jointly intertwined, the (2+1)D blocks (with factorized spatial and temporal components) are easier to optimize. Our experiments demonstrate that ResNets adopting (2+1)D blocks homogeneously in all layers achieve state-of-the-art performance on both Kinetics and Sports-1M.

## 2. Related Work

Video understanding is one of the core computer vision problems and has been studied for decades. Many research contributions in video understanding have focused on developing spatiotemporal features for video analysis. Some proposed video representations include spatiotemporal interest points (STIPs) [21], SIFT-3D [27], HOG3D [18], Motion Boundary Histogram [5], Cuboids [6], and ActionBank [26]. These representations are hand-designed and use different feature encoding schemes such as those based on histograms or pyramids. Among these hand-crafted representations, improved Dense Trajectories (iDT) [38] is widely considered the state-of-the-art, thanks to its strong results on video classification.

After the breakthrough of deep learning in still-image recognition originated by the introduction of the AlexNet model [19], there has been active research devoted to the design of deep networks for video. Many attempts in this genre leverage CNNs trained on images to extract features from the individual frames and then perform temporal integration of such features into a fixed-size descriptor using pooling, high-dimensional feature encoding [41, 11], or recurrent neural networks [42, 7, 32, 2]. Karpathy *et al.* [16] presented a thorough study on how to fuse temporal information in CNNs and proposed a "slow fusion" model that extends the connectivity of all convolutional layers in time and computes activations though temporal convolutions in addition to spatial convolutions. However, they found that the networks operating on individual frames performed on par with the networks processing the entire spatiotemporal volume of the video. 3D CNNs using temporal convolutions for recognizing human actions in video were arguably first proposed by Baccouche *et al.* [1] and by Ji *et al.* [15]. But 3D convolutions were also studied in parallel for unsupervised spatiotemporal feature learning with Restricted Boltzmann Machines [35] and stacked ISA [22]. More recently, 3D CNNs were shown to lead to strong action recognition results when trained on large-scale datasets [36]. 3D CNNs features were also demonstrated to generalize well to other tasks, including action detection [28], video captioning [24], and hand gesture detection [23].

Another influential approach to CNN-based video modeling is represented by the two-stream framework introduced by Simonyan and Zisserman [29], who proposed to fuse deep features extracted from optical flow with the more traditional deep CNN activations computed from color RGB input. Feichtenhofer *et al.* enhanced these two-stream networks with ResNet architectures [13] and additional connections between streams [9]. Additional two-stream approaches include Temporal Segment Networks [39], Action Transformations [40], and Convolutional Fusion [10]. Notably, Carreira and Zisserman recently introduced a model (I3D) that combines two-stream processing and 3D convolutions. I3D currently holds the best action recognition results on the large-scale Kinetics dataset.

Our work revisits many of the aforementioned approaches (specifically 3D CNNs, two-stream networks, and ResNets) in the context of an empirical analysis deeply focused on understanding the effects of different types of spatiotemporal convolutions on action recognition performance. We include in this study 2D convolution over frames, 2D convolution over clips, 3D convolution, interleaved (mixed) 3D-2D convolutions, as well as a decomposition of 3D convolution into a 2D spatial convolution followed by 1D temporal convolution, which we name (2+1)D convolution. We show that when used within a ResNet architecture [13], (2+1)D convolutions lead to state-of-the-art results on **4** different benchmarks in action recognition. Our architecture, called R(2+1)D, is related to Factorized Spatio-Temporal Convolutional Networks [33] ($F_{ST}CN$) in the way of factorizing spatiotemporal convolutions into spatial and temporal ones. However, $F_{ST}CN$ focuses on network factorization, e.g. $F_{ST}CN$ is implemented by several spatial layers at the lower layers and two parallel temporal layers on its top. On the other hand, R(2+1)D focuses on layer factorization, i.e. factorizing each spatiotemporal convolution into a block of a spatial convolution and a temporal convolution. As a result, R(2+1)D is alternating between spatial and temporal convolutions across the network. R(2+1)D is also closely related to the Pseudo-3D network (P3D) [25], which includes three different residual blocks that adapt the bottleneck block of 2D ResNets to video. The blocks implement different forms of spatiotemporal convolution: spatial followed by temporal, spatial and temporal in parallel, and spatial followed by temporal with skip connection from spatial convolution to the output of the block, respectively. The P3D model is formed by interleaving these three blocks in sequence through the depth of the network. In contrast, our R(2+1)D model uses a single type of spatiotemporal residual block homogeneously in all layers and it does not include bottlenecks. Instead, we show that by making a careful choice of dimensionality for the spatial-temporal decomposition in each block we can obtain a model that is compact in size and that yet leads to state-of-the-art action recognition accuracy. For example, on Sports-1M using RGB as input, R(2+1)D outperforms P3D by a margin of $9.1\%$ in Clip@1 accuracy ($57.0\%$ vs $47.9\%$), despite the fact that P3D uses a 152-layer ResNet, while our model has only 34 layers.

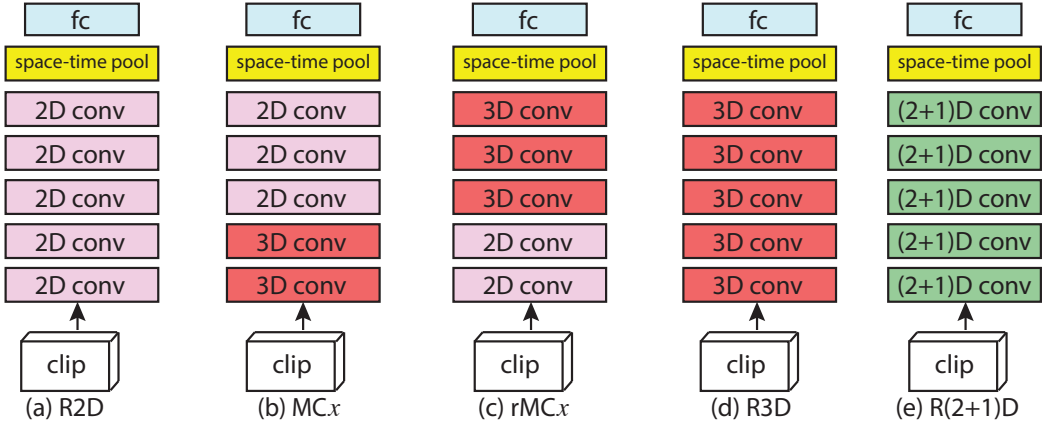

Figure 1. **Residual network architectures for video classification considered in this work**. (a) R2D are 2D ResNets; (b) MCx are ResNets with mixed convolutions (MC3 is presented in this figure); (c) rMCx use reversed mixed convolutions (rMC3 is shown here); (d) R3D are 3D ResNets; and (e) R(2+1)D are ResNets with (2+1)D convolutions. For interpretability, residual connections are omitted.

## 3. Convolutional residual blocks for video

In this section we discuss several spatiotemporal convolutional variants within the framework of residual learning. Let **x** denote the input clip of size $3 \times L \times H \times W$, where $L$ is the number of frames in the clip, $H$ and $W$ are the frame height and width, and 3 refers to the RGB channels. Let $z_i$ be the tensor computed by the $i$-th convolutional block in the residual network. In this work we consider only "vanilla" residual blocks (i.e., without bottlenecks) [13], with each block consisting of two convolutional layers with a ReLU activation function after each layer. Then the output of the $i$-th residual block is given by

$$z_i = z_{i-1} + \mathcal{F}(z_{i-1}; \theta_i) \quad (1)$$

where $\mathcal{F}(; \theta_i)$ implements the composition of two convolutions parameterized by weights $\theta_i$ and the application of the ReLU functions. In this work we consider networks where the sequence of convolutional residual blocks culminates into a top layer performing global average pooling over the entire spatiotemporal volume and a fully-connected layer responsible for the final classification prediction.

### 3.1. R2D: 2D convolutions over the entire clip

2D CNNs for video [29] ignore the temporal ordering in the video and treat the $L$ frames analogously to channels. Thus, we can think of these models as reshaping the input 4D tensor **x** into a 3D tensor of size $3L \times H \times W$. The output $z_i$ of the $i$-th residual block is also a 3D tensor. Its size is $N_i \times H_i \times W_i$ where $N_i$ denotes the number of convolutional filters applied in the $i$-th block, and $H_i, W_i$ are the spatial dimensions, which may be smaller than the original input frame due to pooling or striding. Each filter is 3D and it has size $N_{i-1} \times d \times d$, where $d$ denotes the spatial width and height. Note that although the filter is 3-dimensional, it is convolved only in 2D over the *spatial* dimensions of the preceding tensor $z_{i-1}$. Each filter yields a single-channel output. Thus, the very first convolutional layer in R2D collapses the entire temporal information of the video in single-channel feature maps, which prevent any temporal reasoning to happen in subsequent layers. This type of CNN architecture is illustrated in Figure 1(a). Note that since the feature maps have no temporal meaning, we do not perform temporal striding for this network.

### 3.2. f-R2D: 2D convolutions over frames

Another 2D CNN approach involves processing independently the $L$ frames via a series of 2D convolutional residual block. The same filters are applied to all $L$ frames. In this case, no temporal modeling is performed in the convolutional layers and the global spatiotemporal pooling layer at the top simply fuses the information extracted independently from the $L$ frames. We refer to this architecture variant as f-R2D (frame-based R2D).

### 3.3. R3D: 3D convolutions

3D CNNs [15, 36] preserve temporal information and propagate it through the layers of the network. The tensor $z_i$ is in this case 4D and has size $N_i \times L \times H_i \times W_i$, where $N_i$ is the number of filters used in the $i$-th block. Each filter is 4-dimensional and it has size $N_{i-1} \times t \times d \times d$ where $t$ denotes the temporal extent of the filter (in this work, we use $t = 3$, as in [36, 4]). The filters are convolved in 3D, i.e., over both time and space dimensions. This type of CNN architecture is illustrated in Figure 1(d).

### 3.4. MC$x$ and rMC$x$: mixed 3D-2D convolutions

One hypothesis is that motion modeling (i.e., 3D convolutions) may be particularly useful in the early layers, while at higher levels of semantic abstraction (late layers), motion or temporal modeling is not necessary. Thus a plausible architecture may start with 3D convolutions and switch to using 2D convolutions in the top layers. Since in this work we consider 3D ResNets (R3D) having 5 groups of convolutions (see Table 1), our first variant consists in replacing all 3D convolutions in group 5 with 2D convolutions. We denote this variant with MC5 (Mixed Convolutions). We design a second variant that uses 2D convolutions in group 4 and 5, and name this model MC4 (meaning from group 4 and deeper layers all convolutions are 2D). Following this pattern, we also create MC3 and MC2 variations. We omit to consider MC1 since it is equivalent to a 2D ResNet (f-R2D) applied to clip inputs. This type of CNN architectures is illustrated in Figure 1(b). An alternative hypothesis is that temporal modeling may be more beneficial in the deep layers, with early capturing appearance information via 2D convolutions. To account for such possibility, we also experiment with "Reversed" Mixed Convolutions. Following the naming convention of MC models, we denote these models as rMC2, rMC3, rMC4, and rMC5. Thus, rMC3 would include 2D convolutions in block 1 and 2, and 3D convolutions in group 3 and deeper groups. This type of CNN architecture is illustrated in Figure 1(c).

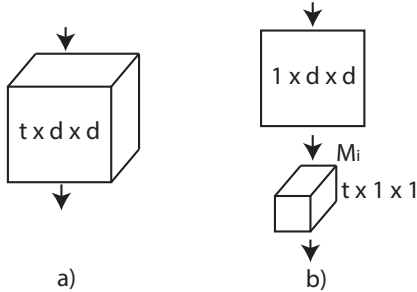
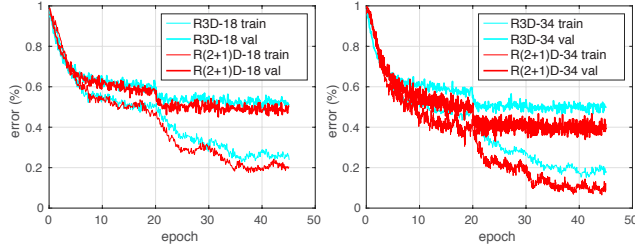

Figure 2. **(2+1)D vs 3D convolution**. The illustration is given for the simplified setting where the input consists of a spatiotemporal volume with a single feature channel. (a) Full 3D convolution is carried out using a filter of size $t \times d \times d$ where $t$ denotes the temporal extent and $d$ is the spatial width and height. (b) A (2+1)D convolutional block splits the computation into a spatial 2D convolution followed by a temporal 1D convolution. We choose the numbers of 2D filters ($M_i$) so that the number of parameters in our (2+1)D block matches that of the full 3D convolutional block.

### 3.5. R(2+1)D: (2+1)D convolutions

Another possible theory is that full 3D convolutions may be more conveniently approximated by a 2D convolution followed by a 1D convolution, decomposing spatial and temporal modeling into two separate steps. We thus design a network architecture named R(2+1)D, where we replace the $N_i$ 3D convolutional filters of size $N_{i-1} \times t \times d \times d$ with a (2+1)D block consisting of $M_i$ 2D convolutional filters of size $N_{i-1} \times 1 \times d \times d$ and $N_i$ temporal convolutional filters of size $M_i \times t \times 1 \times 1$. The hyperparameter $M_i$ determines the dimensionality of the intermediate subspace where the signal is projected between the spatial and the temporal convolutions. We choose $M_i = \lfloor \frac{td^2 N_{i-1} N_i}{d^2 N_{i-1} + t N_i} \rfloor$ so that the number of parameters in the (2+1)D block is approximately equal to that implementing full 3D convolution. We note that this spatiotemporal decomposition can be applied to any 3D convolutional layer. An illustration of this decomposition is given in Figure 2 for the simplified setting where the input tensor $\mathbf{z}_{i-1}$ contains a single channel (i.e., $N_{i-1} = 1$). If the 3D convolution has spatial or temporal striding (implementing downsampling), the striding is correspondingly decomposed into its spatial or temporal dimensions. This architecture is illustrated in Figure 1(e).

Figure 3. **Training and testing errors for R(2+1)D and R3D**. Results are reported for ResNets of 18 layers (left) and 34 layers (right). It can be observed that the training error (thin lines) is smaller for R(2+1)D compared to R3D, particularly for the network with larger depth (right). This suggests that the the spatial-temporal decomposition implemented by R(2+1)D eases the optimization, especially as depth is increased.

Compared to full 3D convolution, our (2+1)D decomposition offers two advantages. First, despite not changing the number of parameters, it doubles the number of nonlinearities in the network due to the additional ReLU between the 2D and 1D convolution in each block. Increasing the number of nonlinearities increases the complexity of functions that can be represented, as also noted in VGG networks [30] which approximate the effect of a big filter by applying multiple smaller filters with additional nonlinearities in between. The second benefit is that forcing the 3D convolution into separate spatial and temporal components renders the optimization easier. This is manifested in lower training error compared to 3D convolutional networks of the same capacity. This is illustrated in Figure 3 which shows training and testing errors for R3D and R(2+1)D having 18 (left) and 34 (right) layers. It can be seen that, for the same number of layers (and parameters), R(2+1)D yields not only lower testing error but also lower training error compared to R3D. This is an indication that optimization becomes easier when spatiotemporal filters are factorized. The gap in the training losses is particularly large for the nets having 34 layers, which suggests that the facilitation in optimization increases as the depth becomes larger.

We note that our factorization is closely related to Pseudo-3D blocks (P3D) [25], which were proposed to adapt the bottleneck block of R2D to video classification. Three different pseudo-3D blocks were introduced: P3D-A,

P3D-B, and P3D-C. The blocks implement different orders of convolution: spatial followed by temporal, spatial and temporal in parallel, and spatial followed by temporal with skip connection from spatial convolution to the output of the block, respectively. Our (2+1)D convolution is most closely related to the P3D-A block, which however contains bottlenecks. Furthermore, the final P3D architecture is composed by interleaving these three blocks in sequence throughout the network, with the exception of the first layer where 2D convolution is used. We propose instead a homogeneous architecture where the same (2+1)-decomposition is used in all blocks. Another difference is that P3D-A is not purposely designed to match the number of parameters with the 3D convolutions. Despite the fact that R(2+1)D is very simple and homogenous in its architecture, our experiments show that it significantly outperforms R3D, R2D, and P3D on Sports-1M (see Table 4).

## 4. Experiments

In this section we present a study of action recognition accuracy for the different spatiotemporal convolutions presented in the previous section. We use Kinetics [4] and Sports-1M [16] as the primary benchmarks, as they are large enough to enable training of deep models from scratch. Since a good video model must also support effective transfer learning to other datasets, we include results obtained by pretraining our models on Sports-1M and Kinetics, and finetuning them on UCF101 [31] and HMDB51 [20].

### 4.1. Experimental setup

**Network architectures.** We constrain our experiments to deep residual networks [13] owing to their good performance and simplicity. Table 1 provides the specifications of the two R3D architectures (3D ResNets) considered in our experiments. The first has 18 layers, while the second variant has 34 layers. Each network takes clips consisting of $L$ RGB frames with the size of $112 \times 112$ as input. We use one spatial downsampling at conv1 implemented by convolutional striding of $1 \times 2 \times 2$, and three spatiotemporal downsampling at conv3_1, conv4_1, and conv5_1 with convolutional striding of $2 \times 2 \times 2$. From these R3D models we obtain architectures R2D, MC$x$, rMC$x$, and R(2+1)D by replacing the 3D convolutions with 2D convolutions, mixed convolutions, reversed mixed convolutions, and (2+1)D convolutions, respectively. Since our spatiotemporal downsampling is implemented by 3D convolutional striding, when 3D convolutions are replaced by 2D ones, e.g., as in MC$x$ and rMC$x$, spatiotemporal downsampling becomes only spatial. This difference yields activation tensors of different temporal sizes in the last convolutional layer. For example, for f-R2D the output of the last convolution layer is $L \times 7 \times 7$, since no temporal striding is applied. Conversely, for R3D and R(2+1)D the last convolutional tensor has size $\frac{L}{8} \times 7 \times 7$. MC$x$ and rMC$x$ models will yield different sizes in the time dimension, depending on how many times temporal striding is applied (as shown in the Table 1). Regardless of the size of output produced by the last convolutional layer, each network applies global spatiotemporal average pooling to the final convolutional tensor, followed by a fully-connected (fc) layer performing the final classification (the output dimension of the fc layer matches the number of classes, e.g., 400 for Kinetics).

| layer name | output size | R3D-18 | R3D-34 |
|---|---|---|---|
| conv1 | L×56×56 | 3×7×7, 64, stride 1 × 2× 2 | |
| conv2_x | L×56×56 | $\begin{bmatrix} 3\times3\times3, 64 \\ 3\times3\times3, 64 \end{bmatrix}\times2$ | $\begin{bmatrix} 3\times3\times3, 64 \\ 3\times3\times3, 64 \end{bmatrix}\times3$ |
| conv3_x | $\frac{L}{2}$×28×28 | $\begin{bmatrix} 3\times3\times3, 128 \\ 3\times3\times3, 128 \end{bmatrix}\times2$ | $\begin{bmatrix} 3\times3\times3, 128 \\ 3\times3\times3, 128 \end{bmatrix}\times4$ |
| conv4_x | $\frac{L}{4}$×14×14 | $\begin{bmatrix} 3\times3\times3, 256 \\ 3\times3\times3, 256 \end{bmatrix}\times2$ | $\begin{bmatrix} 3\times3\times3, 256 \\ 3\times3\times3, 256 \end{bmatrix}\times6$ |
| conv5_x | $\frac{L}{8}$×7×7 | $\begin{bmatrix} 3\times3\times3, 512 \\ 3\times3\times3, 512 \end{bmatrix}\times2$ | $\begin{bmatrix} 3\times3\times3, 512 \\ 3\times3\times3, 512 \end{bmatrix}\times3$ |
| | 1×1×1 | spatiotemporal pooling, fc layer with softmax | |

Table 1. **R3D architectures considered in our experiments**. Convolutional residual blocks are shown in brackets, next to the number of times each block is repeated in the stack. The dimensions given for filters and outputs are time, height, and width, in this order. The series of convolutions culminates with a global spatiotemporal pooling layer that yields a 512-dimensional feature vector. This vector is fed to a fully-connected layer that outputs the class probabilities through a softmax.

**Training and evaluation.** We perform our initial evaluation on Kinetics, using the training split for training and the validation split for testing. For a fair comparison, we set all of the networks to have 18 layers and we train them from scratch on the same input. Video frames are scaled to the size of $128 \times 171$ and then each clip is generated by randomly cropping windows of size $112 \times 112$. We randomly sample $L$ consecutive frames from the video with temporal jittering while training. In this comparison, we experiment with two settings: models are trained on 8-frame clips ($L = 8$) and 16-frame clips ($L = 16$). Batch normalization is applied to all convolutional layers. We use a mini-batch size of 32 clips per GPU. Although Kinetics has only about 240k training videos, we set epoch size to be 1M for temporal jittering. The initial learning rate is set to $0.01$ and divided by 10 every 10 epochs. We use the first 10 epochs for warm-up [12] in our distributed training. Training is done in 45 epochs. We report clip top-1 accuracy and video top-1 accuracy. For video top-1, we use center crops of 10 clips uniformly sampled from the video and average these 10 clip predictions to obtain the video prediction. Training

| Net | # params | Clip@1 | Video@1 | Clip@1 | Video@1 |
|---|---|---|---|---|---|
| Input | | 8×112×112 | | 16×112×112 | |
| R2D | 11.4M | 46.7 | 59.5 | 47.0 | 58.9 |
| f-R2D | 11.4M | 48.1 | 59.4 | 50.3 | 60.5 |
| R3D | 33.4M | 49.4 | 61.8 | 52.5 | 64.2 |
| MC2 | 11.4M | 50.2 | 62.5 | 53.1 | 64.2 |
| MC3 | 11.7M | 50.7 | 62.9 | 53.7 | 64.7 |
| MC4 | 12.7M | 50.5 | 62.5 | 53.7 | 65.1 |
| MC5 | 16.9M | 50.3 | 62.5 | 53.7 | 65.1 |
| rMC2 | 33.3M | 49.8 | 62.1 | 53.1 | 64.9 |
| rMC3 | 33.0M | 49.8 | 62.3 | 53.2 | 65.0 |
| rMC4 | 32.0M | 49.9 | 62.3 | 53.4 | 65.1 |
| rMC5 | 27.9M | 49.4 | 61.2 | 52.1 | 63.1 |
| R(2+1)D | 33.3M | **52.8** | **64.8** | **56.8** | **68.0** |

Table 2. **Action recognition accuracy for different forms of convolution on the Kinetics validation set**. All models are based on a ResNet of 18 layers, and trained from scratch on either 8-frame or 16-frame clip input. R(2+1)D outperforms all the other models.

is done with synchronous distributed SGD on GPU clusters using caffe2 [3].

### 4.2. Comparison of spatiotemporal convolutions

Table 2 reports the clip top-1 and video top-1 action classification accuracy on the Kinetics validation set. There are a few findings that can be inferred from these results. First, there is a noticeable gap between the performance of 2D ResNets (f-R2D and R2D) and that of R3D or mixed convolutional models (MC$x$ and rMC$x$). This gap is $1.3 - 4\%$ in the 8-frame input setting and becomes bigger (i.e. $1.8 - 6.7\%$) when models are trained on 16-frame clips as input. This suggests that motion modeling is important for action recognition. Note that all models (within the same setting) see the same input and process all frames in each clip (either 8 or 16 frames). The difference is that, compared to 3D or MC$x$ models which perform temporal reasoning through the clip, R2D collapses and eliminates temporal information after the first residual block, while f-R2D computes still-image features from the individual frames. Among the different 3D convolutional models, R(2+1)D clearly performs the best. It is $2.1-3.4\%$ better than MC$x$, rMC$x$, R3D in the 8-frame setting, and $3.1-4.7\%$ better in the 16-frame input setting. This indicates that decomposing 3D convolutions in separate spatial and temporal convolutions is better than modeling spatiotemporal information jointly or via mixed 3D-2D convolutions. It also outperforms 2D ResNets (R2D and f-R2D) by $4.7-6.1\%$ in the 8-frame setting and by $6.3-9.8\%$ in the 16-frame input setting.

Figure 4 shows video top-1 accuracy on Kinetics validation set versus computational complexity (FLOPs) for different models. Figure 4(a) plots the models trained on 8-frame clips while Figure 4(b) shows models with 16-frame clip input. The most efficient network is R2D but it has the poorest accuracy. In fact, R2D is about 7x faster than f-R2D because it collapses the temporal dimension after `conv1`.

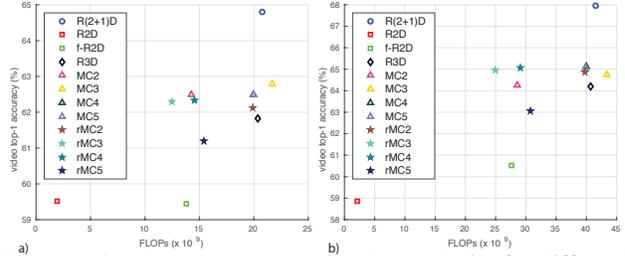

Figure 4. **Accuracy vs computational complexity for different types of convolution on Kinetics**. Different models are trained on 8-frame clips (left) and 16-frame clips (right). R(2+1)D achieves the highest accuracy, producing about $3-3.8\%$ accuracy gain over R3D for the same computational cost.

In terms of accuracy, R2D gets similar performance to f-R2D when trained on 8-frame clips, while it is $1.6\%$ worse than f-R2D in the 16-frame input setting. This is because R2D performs temporal modeling only in the `conv1` layer and thus it handles poorly longer clip inputs. Interestingly, rMC3 is more efficient than f-R2D since it performs temporal striding in `conv3_1`, which yields smaller activation tensors in all subsequent 2D convolutional layers. Conversely, f-R2D processes all frames independently and does not perform any temporal striding. rMC2 is more costly than rMC3, as it uses 2D convolutions in group 2, and does not perform temporal striding in group 3. R(2+1)D has roughly the same computational cost as R3D but it yields higher accuracy. We note that the relative ranking between different architectures is consistent across the two input settings (8 vs 16 frame-clips). However, the gaps are bigger for the 16-frame input setting. This indicates that temporal modeling is more beneficial on longer clip inputs.

**Why are (2+1)D convolutions better than 3D?** Figure 3 presents the training and testing errors on Kinetics for R3D and R(2+1)D, using 18-layers (left) and 34 layers (right). We already know that R(2+1)D gives lower testing error than R3D but the interesting message in this plot is that R(2+1)D yields also lower *training* error. The reduction in training error for R(2+1)D compared to R3D is particularly accentuated for the architecture having 34 layers. This suggests that the spatiotemporal decomposition of R(2+1)D renders the optimization easier compared to R3D, especially as depth is increased.

### 4.3. Revisiting practices for video-level prediction

Varol *et. al.* [37] showed that accuracy gains can be obtained by training video CNNs on longer input clips (e.g. with 100 frames) using long-term convolutions (LTC). Here we revisit this idea and evaluate this practice on Kinetics using R(2+1)D of 18 layers with varying input clip lengths: 8, 16, 24, 32, 40, and 48 frames. The outputs of the last convolution layer for these networks have different temporal sizes, but once again we use a global spatiotemporal average pooling to generate a fixed-size representation which is fed to the fully-connected layer. Note that these networks have the same number of parameters (since pooling involves no learnable parameters). They simply see inputs of different lengths. Figure 5 a) plots the clip-level and video-level accuracy on Kinetics validation set with respect to different input lengths. Note that video-level prediction is done by averaging the clip-level predictions obtained for 10 clips evenly spaced in the video. One interesting finding is that, although clip accuracy continues to increase when we add more frames, video accuracy peaks at 32 frames.

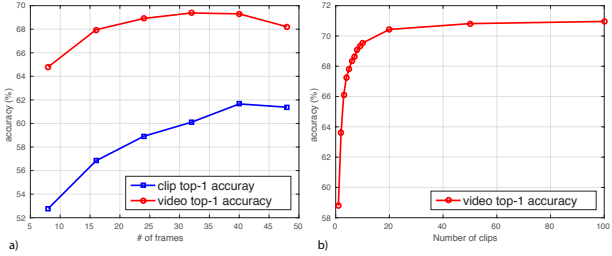

Figure 5. **Video-level accuracy on Kinetics**. a) Clip and video accuracy of a 18-layer R(2+1)D trained on clips of different lengths. b) Video top-1 accuracy obtained by averaging over different number of clip predictions using the same model with 32 frames.

| train | finetune | test | training time | Clip@1 | Video@1 |
|---|---|---|---|---|---|
| clip length (in frames) | | | hours | (%) | (%) |
| 8 | none | 8 | 11.8 | 52.8 | 64.8 |
| 8 | none | 32 | 11.8 | 51.6 | 59.0 |
| 32 | none | 32 | 59.8 | 60.1 | 69.4 |
| 8 | 32 | 32 | 20.5 | 59.8 | 68.0 |

Table 3. **Training time/accuracy trade-off**. Training and evaluation of R(2+1)D with 18 layers on clips with different length. Training on shorter clips, then finetuning on longer clips gives the best trade-off between total training time and accuracy. The training time is measured with 64 GPUs training in parallel.

| method | Clip@1 | Video@1 | Video@5 |
|---|---|---|---|
| DeepVideo [16] | 41.9 | 60.9 | 80.2 |
| C3D [36] | 46.1 | 61.1 | 85.2 |
| 2D Resnet-152 [13] | 46.5* | 64.6* | 86.4* |
| Conv pooling [42] | - | 71.7 | 90.4 |
| P3D [25] | 47.9* | 66.4* | 87.4* |
| R3D-RGB-8frame | 53.8 | - | - |
| R(2+1)D-RGB-8frame | 56.1 | 72.0 | 91.2 |
| R(2+1)D-Flow-8frame | 44.5 | 65.5 | 87.2 |
| R(2+1)D-Two-Stream-8frame | - | 72.2 | 91.4 |
| R(2+1)D-RGB-32frame | **57.0** | 73.0 | 91.5 |
| R(2+1)D-Flow-32frame | 46.4 | 68.4 | 88.7 |
| R(2+1)D-Two-Stream-32frame | - | **73.3** | **91.9** |

Table 4. **Comparison with the state-of-the-art on Sports-1M**. R(2+1)D outperforms C3D by 10.9%, and P3D by 9.1% and it achieves the best reported accuracy on this benchmark to date. *These baseline numbers are taken from [25].

Since all these models have the same numbers of parameters, it is natural to ask "what causes the differences in video-level accuracies?" In order to address this question, we conduct two experiments. In the first experiment, we take the model trained on 8-frame clips and test it using 32-frame clips as input. We found that this causes a drop of 1.2% in clip accuracy and 5.8% in video accuracy compared to testing on 8-frame clips. In the second experiment, we finetuned the 32-frame model using as initialization the parameters of the 8-frame model. In this case, the net achieves results that are almost as good as when learning from scratch on 32-frame clips (59.8% vs 60.1%) and produces a gain of 7% over the 8-frame model. The advantage, however, is that finetuning the 32-frame model from the 8-frame net shortens considerably the total training time versus learning from scratch, since the 8-frame model is 7.3x faster than the 32-frame model in terms of FLOPs. These two experiments suggest that training on longer clips yields different (better) clip-level models, as the filters learn longer-term temporal patterns. This improve-ment cannot be obtained "for free" by simply lengthening the clip input at test time. This is consistent with the findings in [37]. Table 3 reports the total training time and accuracy of R(2+1)D with 18 layers trained and evaluated on clips of varying length.

**How many clips are needed for accurate video-level prediction?** Figure 5 b) plots the video top-1 accuracy of R(2+1)D with 18 layers trained on 32-frame clips when we vary the number of clips sampled from each video. Using 20 crops is only about 0.5% worse than using 100 crops, but the prediction is 5x faster.

### 4.4. Action recognition with a 34-layer R(2+1)D net

In this section we report results using a 34-layer version of R(2+1)D, which we denote as R(2+1)D-34. The architecture is the same as that shown in the right column of Table 1, but with 3D convolutions decomposed spatiotemporally in (2+1)D. We train our R(2+1)D architecture on both RGB and optical flow inputs and fuse the prediction scores by averaging, as proposed in the two-stream framework [29] and subsequent work [4, 9]. We use Farneback's method [8] to compute optical flow because of its efficiency.

**Datasets**. We evaluate our proposed R(2+1)D architecture on four public benchmarks. *Sports-1M* is a large-scale dataset for classification of sport videos [16]. It includes 1.1M videos of 487 fine-grained sport categories. It is provided with a train/test split. *Kinetics* has about 300K videos of 400 human actions. We report results on the validation set as the annotations on the testing set is not public available. *UCF101* and *HMDB51* are well-established benchmarks for action recognitions. UCF101 has about 13K videos of 101 categories, whereas HMDB51 is slightly smaller with 6K videos of 51 classes. Both UCF101 and HMDB51 are provided with 3 splits for training and testing. We report the accuracy by averaging over all 3 splits.

**Results on Sports-1M.** We train R(2+1)D-34 on Sports-1M [16] with both 8-frame and 32-frame clip inputs. Train-ing is done with a setup similar to that described in section 4.1. We also train a R3D-34 baseline for comparison. Videos in Sports-1M are very long, over 5 minutes on average. Thus, we uniformly sample 100 clips per video (instead of 10 clips on Kinetics) for computing the video top-1 accuracy. Average pooling is used to aggregate the predictions over the 100 clips to obtain the video-level predictions.

| method | pretraining dataset | top1 | top5 |
|---|---|---|---|
| I3D-RGB [4] | none | 67.5 | 87.2 |
| I3D-RGB [4] | ImageNet | 72.1 | 90.3 |
| I3D-Flow [4] | ImageNet | 65.3 | 86.2 |
| I3D-Two-Stream [4] | ImageNet | **75.7** | **92.0** |
| R(2+1)D-RGB | none | 72.0 | 90.0 |
| R(2+1)D-Flow | none | 67.5 | 87.2 |
| R(2+1)D-Two-Stream | none | 73.9 | 90.9 |
| R(2+1)D-RGB | Sports-1M | 74.3 | 91.4 |
| R(2+1)D-Flow | Sports-1M | 68.5 | 88.1 |
| R(2+1)D-Two-Stream | Sports-1M | **75.4** | **91.9** |

Table 5. **Comparison with the state-of-the-art on Kinetics**. R(2+1)D outperforms I3D by 4.5% when trained from scratch on RGB. R(2+1)D pretrained on Sports-1M outperforms I3D pretrained on ImageNet, for both RGB and optical flow. However, it is slightly worse than I3D (0.3%) when fusing the two streams.

| method | pretraining dataset | UCF101 | HMDB51 |
|---|---|---|---|
| Two-Stream [29] | ImageNet | 88.0 | 59.4 |
| Action Transf. [40] | ImageNet | 92.4 | 62.0 |
| Conv Pooling [42] | Sports-1M | 88.6 | - |
| $F_{ST}CN$ [33] | ImageNet | 88.1 | 59.1 |
| Two-Stream Fusion [10] | ImageNet | 92.5 | 65.4 |
| Spatiotemp. ResNet [9] | ImageNet | 93.4 | 66.4 |
| Temp. Segm. Net [39] | ImageNet | 94.2 | 69.4 |
| P3D [25] | ImageNet+Sports1M | 88.6 | - |
| I3D-RGB [4] | ImageNet+Kinetics | 95.6 | 74.8 |
| I3D-Flow [4] | ImageNet+Kinetics | 96.7 | 77.1 |
| I3D-Two-Stream [4] | ImageNet+Kinetics | **98.0** | **80.7** |
| R(2+1)D-RGB | Sports1M | 93.6 | 66.6 |
| R(2+1)D-Flow | Sports1M | 93.3 | 70.1 |
| R(2+1)D-TwoStream | Sports1M | 95.0 | 72.7 |
| R(2+1)D-RGB | Kinetics | 96.8 | 74.5 |
| R(2+1)D-Flow | Kinetics | 95.5 | 76.4 |
| R(2+1)D-TwoStream | Kinetics | 97.3 | 78.7 |

Table 6. **Comparison with the state-of-the-art on UCF101 and HMDB51**. Our R(2+1)D finetuned from Kinetics is nearly on par with I3D which, however, uses ImageNet in addition to Kinetics for pretraining. We found that Kinetics is better than Sports-1M for pretraining our models.

Table 4 shows the results on Sports-1M. Our R(2+1)D model trained on RGB performs the best among the methods in this comparison. In clip-level accuracy, it outperforms C3D by 10.9% and P3D by 9.1%. R(2+1)D also outperforms 2D ResNet by 10.5%. We note that the 2D ResNet and P3D have 152 layers while R(2+1)D has only 34 layers (or 67 if we count the spatiotemporal decomposition as producing two layers). The R3D baseline is also inferior to R(2+1)D (by 2.3%) when the input is 8 RGB frames, which confirms the benefits of our (2+1)D decomposition. R(2+1)D achieves a video-level accuracy of 73.3% which, to our knowledge, is the best published result on Sports-1M.

**Results on Kinetics.** We assess the performance of R(2+1)D-34 on Kinetics, both when training from scratch on the Kinetics training split, as well as when finetuning the model pretrained on Sports-1M. When training from scratch, we use the same setup as in section 4.1. When finetuning, we start with a base learning rate that is 10 times smaller (i.e., 0.001), and reduce it by a factor of 10 every 4 epochs. Finetuning is completed at 15 epochs. Table 5 reports the results on Kinetics. R(2+1)D outperforms I3D by 4.5% when both models are trained from scratch on RGB input. This indicates that our R(2+1)D is a competitive architecture for action recognition. Our R(2+1)D pretrained on Sports-1M also outperforms I3D pretrained on ImageNet by 2.2% when using RGB as input and by 3.2% when trained on optical flow. However, it is slightly worse than I3D (by 0.3%) when fusing the two streams.

**Transferring models to UCF101 and HMDB51.** We also experiment with finetuning R(2+1)D on UCF101 [31] and HMDB51 [20] using models pretrained on Sports-1M and

Kinetics. For the models based on Kinetics pretraining, we use the models learned from scratch on Kinetics (not those finetuned from Sports-1M) in order to understand the effects of pretraining on different datasets. Table 6 reports results of R(2+1)D compared to prior methods. R(2+1)D outperforms all methods in this comparison, except for I3D which, however, used ImageNet in addition to Kinetics for pretraining. R(2+1)D (with Kinetics pretraining) is comparable to I3D when trained on RGB but it is slightly worse than I3D when trained on optical flow. This can be explained by noting that R(2+1)D uses Farneback's optical flow, while I3D uses a more accurate flow, TV-L1 [43] which is, however, one order of magnitude slower than Farneback's method.

## 5. Conclusions

We have presented an empirical study of the effects of different spatiotemporal convolutions for action recognition in video. Our proposed architecture R(2+1)D achieves results comparable or superior to the state of the art on Sports-1M, Kinetics, UCF101, and HMDB51. We hope that our analysis will inspire new network designs harnessing the potential efficacy and modeling flexibility of spatiotemporal convolutions. While our study was focused on a single type of network (ResNet) and a homogenous use of our (2+1)D spatiotemporal decomposition, future work will be devoted to searching more suitable architectures for our approach.

**Acknowledgements**. The authors would like to thank Ahmed Taei, Aarti Basant, Aapo Kyrola, and the Facebook Caffe2 team for their help in implementing ND-convolution, in optimizing video I/O, and in providing support for distributed training. We are grateful to Joao Carreira for sharing I3D results on the Kinetics validation set.

# Appendix

Figure 6 illustrates the decomposed filters of our R(2+1)D model at the `conv1` layer. We remind the readers that, instead of using $64$ 3D filters of size $3 \times 7 \times 7$ at `conv1` as in R3D, R(2+1)D decomposes `conv1` into $45$ 2D filters of size $1 \times 7 \times 7$ and $64$ 1D filters of size $3 \times 1 \times 1$ with non-linear ReLUs in between. This (2+1)D convolutional block has the same number of parameters as that of R3D. Figure 6(a) shows the $45$ spatial filters of size $7 \times 7$ upscaled by 4x for better visualization. Figure 6(b) presents the $64$ temporal filters from left to right. Each temporal filter is visualized as a $45 \times 3$ matrix. Each matrix shows how the temporal filter combines the $45$ channels from the spatial filters across time (3 frames). Some interesting temporal patterns can be seen by looking at the filter weights along the time dimension.

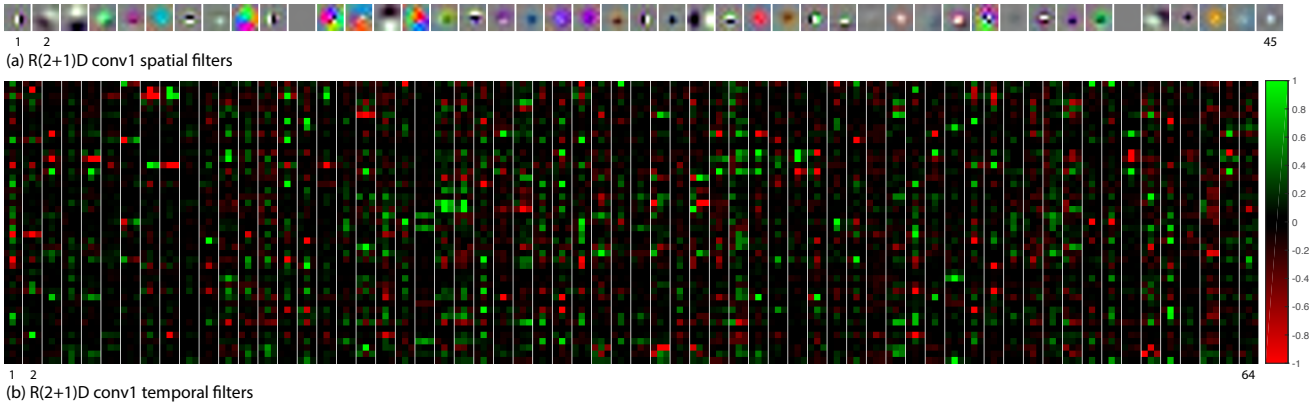

Figure 6. **R(2+1)D learned filters**. (a) The 45 spatial 2D filters learned by R(2+1)D at the `conv1` layer. Each filter has size $1 \times 7 \times 7$ (upscaled by 4x for better visualization). (b) The 64 temporal 1D filters learned by R(2+1)D at the `conv1` layer. Each temporal filter is visualized as a $45 \times 3$ matrix (one column in (b)). The filter weight magnitudes are scaled from $[-0.3, 0.3]$ to $[-1, 1]$ for better presentation. Best viewed in color.